\documentclass[letterpaper, 10 pt, conference]{ieeeconf}

\IEEEoverridecommandlockouts                              

\usepackage{hyperref}
\usepackage{url}
\usepackage{bbold}
\usepackage{amsfonts}
\usepackage{amsmath,amssymb}
\usepackage{slashbox}

\usepackage{graphicx}
\usepackage{hyperref}
\usepackage{wrapfig}
\usepackage{mathrsfs} 
\usepackage{algorithm,algorithmic}
\usepackage{array}
\usepackage{times}
\usepackage{url}
\usepackage{cite}
\usepackage{upgreek}
\usepackage{bbm}
\usepackage{float}
\usepackage{color}

\usepackage{pgfplots}
\usepackage{color}
\usepackage{flushend}
\usepackage{multirow}
\usepackage{rotating}
\usepackage{tikz,pgfplots}
\usepackage{standalone} 
\usepackage{amssymb}

\newcolumntype{M}[1]{>{\centering\arraybackslash}m{#1}}

\newcolumntype{C}[1]{>{\centering\let\newline\\\arraybackslash\hspace{0pt}}m{#1}}

\newcommand\numberthis{\addtocounter{equation}{1}\tag{\theequation}}
\newcolumntype{M}[1]{>{\centering\arraybackslash}m{#1}}

\newcommand{\E}{\mathbb{E}}
\newcommand{\m}{\mathbf{m}}
\newcommand{\J}{\mathbf{J}}
\newcommand{\g}{\mathbf{A}}
\newcommand{\ba}{\mathbf{U}}
\newcommand{\oo}{\mathcal{O}}
\newcommand{\mh}{\widehat{\boldsymbol \mu}}
\newcommand{\dha}{\widehat{\boldsymbol \Delta}}

\newcommand{\sr}{{\tt SAR}}
\newcommand{\ns}{{\tt NSAR}}
\newcommand{\ucb}{{\tt AT-LUCB}}
\newcommand{\uni}{{\tt UNI}}

\newcommand{\hone}{H_1^{\left<M\right>}}
\newcommand{\htwo}{H_2^{\left<M\right>}}
\newcommand{\hp}{H^{\left<M\right>}(p)}

\providecommand{\abs}[1]{\left\vert#1\right\vert}

\providecommand{\seal}[1]{\left\lceil#1\right\rceil}
\providecommand{\p}[1]{\mathbb{P}\left(#1\right)}
\providecommand{\ex}[1]{\exp\left(#1\right)}
\providecommand{\cl}[1]{\colon\hspace{-0.1cm}#1}

\newtheorem{theorem}{Theorem}
\newtheorem{corollary}[theorem]{Corollary}

\newtheorem{problem}[theorem]{Problem}
\newtheorem{proposition}[theorem]{Proposition}

\newtheorem{fact}{Fact}

\begin{document}

\title{\LARGE \bf  Nonlinear Sequential Accepts and Rejects for Identification\\ of Top Arms in Stochastic Bandits}

%

\author{Shahin Shahrampour and Vahid Tarokh
\thanks{This work was supported by DARPA under grant number N6600115C4028.}
\thanks{S. Shahrampour and V. Tarokh are with the John A. Paulson School of Engineering and Applied Sciences, Harvard University, Cambridge, MA, 02138 USA. (e-mail: {\tt \{shahin,vahid\}@seas.harvard.edu}).}
}

\maketitle

\begin{abstract}
We address the $M$-best-arm identification problem in multi-armed bandits. A player has a limited budget to explore $K$ arms ($M<K$), and once pulled, each arm yields a reward drawn (independently) from a fixed, unknown distribution. The goal is to find the top $M$ arms in the sense of expected reward. We develop an algorithm which proceeds in rounds to deactivate arms iteratively. At each round, the budget is divided by a nonlinear function of remaining arms, and the arms are pulled correspondingly. Based on a decision rule, the deactivated arm at each round may be accepted or rejected. The algorithm outputs the accepted arms that should ideally be the top $M$ arms. 
We characterize the decay rate of the misidentification probability and establish that the nonlinear budget allocation proves to be useful for different problem environments (described by the number of competitive arms). We provide comprehensive numerical experiments showing  that our algorithm outperforms the state-of-the-art using suitable nonlinearity. 
\end{abstract}


\section{Introduction}
Multi-Armed Bandits (MAB) is a sequential decision-making framework for the exploration-exploitation dilemma \cite{lai1985asymptotically,auer2002finite}. In MAB, a player explores a finite set of arms, and pulling each arm reveals a {\it reward} to the player. In the stochastic MAB, the rewards for each arm are independent samples from an {\it unknown}, fixed distribution. The player aims to exploit the arm with the largest expected reward as often as possible to maximize the gain. This framework has been formulated in terms of the {\it cumulative} regret, a comparison measure between the player's performance versus a clairvoyant knowing the best arm \emph{a priori}. Early studies on MAB dates back to several decades ago, but the problem has attracted a lot of renewed interest due to its modern applications, such as web
search and advertising, wireless cognitive radios, and multi-channel communication systems (see e.g. \cite{mahajan2008multi,liu2010distributed,wang2012optimality,vakili2013deterministic,kalathil2014decentralized} and references therein).

More recently, many researchers have examined MAB in a pure-exploration framework where the player aims to minimize the {\it simple} regret. This task is closely related to (probability of) finding the best arm in the pool \cite{audibert2010best}. As a result, the best-arm identification problem has received a considerable attention in the literature of machine learning\cite{even2002pac,mannor2004sample,bubeck2009pure,bubeck2011pure,audibert2010best,karnin2013almost,gabillon2012best}. It is well-known that algorithms developed to minimize the cumulative regret (exploration-exploitation) perform poorly for the simple-regret minimization (pure-exploration). Consequently, one must adopt different strategies for optimal best-arm recommendation \cite{bubeck2011pure}. To motivate the pure-exploration setting, consider channel allocation for mobile phone communication. Before the outset of communication, a cellphone (player) can explore the set of channels (arms) to find the best one to operate. Each channel feedback is noisy, and the number of trials (budget) is limited. The problem is hence an instance of best-arm identification, and minimizing the cumulative regret is not the right approach to the problem \cite{audibert2010best}. 

In this paper, we consider the $M$-best-arm identification problem in the {\it fixed-budget} setting \cite{bubeck2013multiple}. Given a fixed number of arm pulls, the player attempts to maximize the probability of correctly identifying the top $M$ arms (in the sense of the expected reward). Note that this setting differs from the {\it fixed-confidence} setting, in which the objective is to minimize the number of trials to find the top $M$ arms with a certain confidence \cite{kalyanakrishnan2010efficient,kalyanakrishnan2012pac}. Recently, for best-arm identification ($M=1$) in the fixed-budget setting, the authors of \cite{shahinjournal} proposed an efficient algorithm based on nonlinear sequential elimination. The idea is to discard the suboptimal arms sequentially and divide the budget according to a \emph{nonlinear function} of remaining arms at each round. With a suitable nonlinearity, the nonlinear budget allocation was proven to improve upon Successive Rejects \cite{audibert2010best} (its linear counterpart) as well as Sequential Halving \cite{karnin2013almost}.

Inspired by the success of nonlinear budget allocation for best-arm identification \cite{shahinjournal}, in this work, we extend the Successive Accepts and Rejects (\sr) algorithm in \cite{bubeck2013multiple} to nonlinear budget allocation for $M$-best-arm identification. Our algorithm, called Nonlinear Sequential Accepts and Rejects (\ns), proceeds in rounds. At each round, the arms are pulled strategically and their empirical rewards are calculated. Then, one arm is deactivated, and according to a decision rule the arm may be accepted or rejected. Unlike \sr~that divides the budget by a linear function of remaining arms, \ns~(our algorithm) does so in a nonlinear fashion. For two general reward regimes, we prove theoretically that our algorithm achieves a lower sample complexity compared to \sr, which improves the decay rate of the misidentification probability. We also provide various numerical experiments to support our theoretical results, and moreover, we compare \ns~to the fixed-budget version of \ucb~in \cite{jun2016anytime}.

\subsection{Related Work} 
Pure-exploration in the PAC-learning setup was examined in \cite{even2002pac}, where Successive Elimination for finding an $\epsilon$-optimal arm with probability $1-\delta$ (fixed-confidence setting) was developed. The matching lower bounds for the problem were provided in \cite{mannor2004sample,even2006action}. Many algorithms for pure-exploration are inspired by the celebrated {\tt UCB1} algorithm for exploration-exploitation \cite{auer2002finite}. As an example, Audibert et al. \cite{audibert2010best} proposed {\tt UCB-E}, which modifies {\tt UCB1} for pure-exploration. In addition, Jamieson et al. \cite{jamieson2014lil} proposed an optimal algorithm for the fixed-confidence setting, inspired by the law of the iterated logarithm. Gabillon et al. \cite{gabillon2012best} presented a unifying approach for fixed-budget and fixed-confidence settings. For identification of multiple top arms (or $M$-best-arm identification), Kalyanakrishnan et al. \cite{kalyanakrishnan2010efficient} developed the {\tt HALVING} algorithm in the fixed-confidence setting, which is later improved by the {\tt LUCB} algorithm in\cite{kalyanakrishnan2012pac}. For the fixed-confidence setting, more recent progress can be found in \cite{chen2017nearly,jiang2017practical,chen2017adaptive}. In \cite{zhou2014optimal}, the $M$-best-arm identification problem was posed using a notion of aggregate regret, and it was applied to crowdsourcing. Furthermore, Kaufmann et al. \cite{kaufmann2013information} studied the identification of multiple top arms using KL-divergence-based confidence intervals. The authors of \cite{kaufmann2016complexity} investigated both settings to show that the complexity of the fixed-budget setting may be smaller than that of the fixed-confidence setting. 


\section{Preliminaries}\label{Preliminaries}
{\bf Notation:} For integer $K$, we define $[K]:=\{1,\ldots,K\}$ to represent the set of positive integers smaller than or equal to $K$. We use $\abs{S}$ to denote the cardinality of the set $S$, and $\seal{\cdot}$ to denote the ceiling function, respectively. We use the notation $f(x)=\oo(g(x))$ when there exists a positive constant $L>0$ and a point $x_0$ such that $\abs{f(x)}\leq L\abs{g(x)}$ for $x \geq x_0$. Throughout, the random variables are denoted in bold letters.

\begin{table*}[t!]
\caption{The parameters $\alpha$ and $\beta$ for the algorithms proposed for (single) best-arm identification. The misidentification probability for each algorithm decays in the form of $\beta \ex{- T/\alpha}$. The quantities used in the table are defined in \eqref{H2} and \eqref{HP}.}
\begin{center}
\resizebox{14cm}{!}{
\begin{tabular}{| c | M{4.3cm} | M{4.3cm} | M{4.3cm} | @{}m{0pt}@{} |} 
 \hline 
  Algorithm & Successive Rejects  & Sequential Halving &  Nonlinear Sequential Elimination &  \\ [.5 ex]
 \hline 
 $\alpha$ & $H_2\overline{\log} K$ & $8H_2 \log_2K$ & $H(p)C_p$ &\\ [.5 ex]
 \hline
 $\beta$ & $0.5K(K-1) \ex{K/(H_2\overline{\log} K)}$ & $3 \log_2K$ & $(K-1)\ex{K/H(p)C_p}$ & \\ [.5 ex]
 \hline
\end{tabular}}
\end{center}\label{table}
\end{table*}

\begin{table*}[t!]
\caption{The sampling complexity of algorithms proposed for $M$-best-arm identification. It identifies the smallest $T$ for which each algorithm recommends the top $M$ arms with probability at least $1-\delta$. The quantities used in the table are defined in \eqref{HP} and \eqref{H2M}.}
\begin{center}
\begin{tabular}{| c | M{3.3cm} | M{3.3cm} | M{3.3cm} | @{}m{0pt}@{} |} 
 \hline 
 Algorithm & \sr  & \ucb &  \ns~(our algorithm) &  \\ [.5 ex]
 \hline 
 Sampling complexity order & $\htwo \overline{\log} K \log\frac{K}{\delta}$ & $\hone \log\frac{\hone}{\delta}$ & $\hp C_p \log\frac{K}{\delta}$ &\\ [.5 ex]
 \hline
\end{tabular}
\end{center}\label{table20}
\end{table*}

\subsection{Problem Statement}
In the stochastic Multi-armed Bandit (MAB) problem, a player explores a finite set of $K$ arms. When the player samples an arm, the corresponding {\it reward} of that arm is observed. The rewards of each arm $i\in [K]$ are drawn independently from an {\it unknown, fixed} distribution with the expected value $\mu_i$. The support of the distribution is the unit interval $[0,1]$, and the rewards are generated independently across the arms. For simplicity, we have the following assumption on the order of arms 
\begin{align}\label{order}
\mu_1 > \mu_2  > \cdots > \mu_K,
\end{align}
where the strict inequalities guarantee that there is no ambiguity over the top $M$ arms $[M]$. Let $\Delta_i:=\mu_1-\mu_i$ denote the {\it gap} between arm $i$ and arm 1, measuring the sub-optimality of arm $i$, and $\mh_{i,n}$ the (empirical) average reward obtained by pulling arm $i$ for $n$ times. 

In this work, we address the $M$-best-arm identification setup, a pure-exploration problem in which the player aims to find the top $M$ arms $[M]$ with a high probability. The two well-known settings for this problem are the fixed-confidence and the fixed-budget. In the former, the objective is to minimize the number of arm pulls needed to identify the top $M$ arms with a certain confidence. In the latter, which is the focus of this work, the problem is posed formally as: 
\begin{problem}
{\it Given a total budget of $T$ arm pulls, an $M$-best-arm identification algorithm outputs the arms $\{\J_1,\ldots,\J_M\}$. Find the decay rate of misidentification probability, i.e., the decay rate of $\p{\{\J_1,\ldots,\J_M\} \neq [M]}$. }
\end{problem}
For the case that $M=1$, known as best-arm identification, it is proven that classical MAB techniques in the exploration-exploitation setting (e.g. {\tt UCB1}) are not optimal. In particular, Bubeck et al. \cite{bubeck2011pure} have showed that upper bounds on the cumulative regret results in lower bounds on the simple regret, i.e., the smaller the cumulative regret, the larger the simple regret. 
The underlying intuition is that in the exploration-exploitation setting, we aim to find the best arm {\it as quickly as possible} to exploit it, and in this case, playing even the second-best arm for a long time yields an unacceptable cumulative regret. On the other hand, in the best-arm identification problem, there is no need to minimize an intermediate cost, and the player only recommends the best arm at the end. Therefore, exploring the suboptimal arms {\it strategically} during the game helps the player to make a better final decision. In other words, the performance is only measured by the final output, regardless of the number of pulls for the suboptimal arms.

\subsection{Previous Performance Guarantees and Our Result}
Though the focus of this work is $M$-best-arm identification, we start by reviewing some of the results for the case of $M=1$ (best-arm identification). Any (single) best-arm identification algorithm samples the arms based on some strategy and outputs a single arm as the best. In order to characterize the misidentification probability of these algorithms, we need to define a few quantities. The decay rate of misidentification probability for two of the state-of-the-art algorithms, Successive Rejects \cite{audibert2010best} and Sequential Halving \cite{karnin2013almost}, relies on the complexity measure $H_2$, defined as 
\begin{align}\label{H2}
H_1:=\sum_{i=2}^K \frac{1}{\Delta_i^2}  \ \ \ \ \  \ \ \  \text{and}  \ \ \ \ \ \  \ \ H_2:=\max_{i \neq 1} \frac{i}{\Delta_i^2},
\end{align}
which is equal to $H_1$ up to logarithmic factors in $K$ \cite{audibert2010best}. In Successive Rejects, at round $r$, the $K-r+1$ remaining arms are played proportional to the whole budget divided by $K-r+1$ (a linear function of $r$). As the linear function is not necessarily the best sampling rule, the authors of \cite{shahinjournal} extended Successive Rejects to Nonlinear Sequential Elimination which divides the budget at round $r$ by the nonlinear function $(K-r+1)^p$, based on an input parameter $p\in (0,2]$ ($p=1$ recovers Successive Rejects). The performance of the algorithm depends on the following quantities
\begin{align}\label{HP}
H(p):=\max_{i \neq 1} \frac{i^p}{\Delta_i^2} \  \ \ \  \ \  \text{and}   \ \ \ \  \ \ C_p:=2^{-p}+\sum_{r=2}^{K}r^{-p}.
\end{align}
For each of the three algorithms, the bound on the misidentification probability can be written in the form of $\beta \ex{- T/\alpha}$, where $\alpha$ and $\beta$ are provided in Table \ref{table} ($\overline{\log}~K=0.5+\sum_{i=2}^Ki^{-1}$). It was shown in \cite{shahinjournal} that in many regimes for the arm gaps, $p\neq 1$ provides better results (theoretical and practical), and Nonlinear Sequential Elimination outperforms the other two algorithms. The value of $p$ must be tuned, but the tuning is more qualitative rather than quantitative, i.e., the algorithm performs reasonably well as long as $p$ is either in $(0,1)$ or $(1,2)$, and thus, the value of $p$ needs not be specific.

In this work, our goal is to extend this idea to $M$-best-arm identification. For convenience, we discuss the performance of these algorithms in terms of the {\it sample complexity}, defined as the smallest budget $T$ needed to achieve the confidence level $\delta$ for misidentification probability, i.e., the smallest $T$ for which $\p{\{\J_1,\ldots,\J_M\} \neq [M]}\leq \delta$. For $M$-best-arm identification, we need to define a new set of quantities and complexity measures as
\begin{align*}\label{H2M}
\Delta_i^{\left<M\right>}&= 
\begin{cases}
    \mu_i-\mu_{M+1},& \text{if } i \leq M\\
\mu_M-\mu_i,              & \text{otherwise}
\end{cases}\\
\hone&=\sum_{i=1}^K\left(\Delta_i^{\left<M\right>}\right)^{-2}\\
\htwo&=\max_{i\neq 1}\left\{i\left(\Delta_{(i)}^{\left<M\right>}\right)^{-2}\right\}\\
\hp&=\max_{i\neq 1}\left\{i^p\left(\Delta_{(i)}^{\left<M\right>}\right)^{-2}\right\}, \numberthis
\end{align*}
where $\Delta_{(i)}^{\left<M\right>}$ for each $(i)\in [K]$ is such that $$\Delta_{(1)}^{\left<M\right>} \leq \Delta_{(2)}^{\left<M\right>} \leq \cdots \leq \Delta_{(K)}^{\left<M\right>}.$$
Based on the definitions above, $$H^{\left<M\right>}(1)=\htwo \neq \hone.$$ Table \ref{table20} tabulates the sample complexities of three algorithms for $M$-best-arm identification: \sr~\cite{bubeck2013multiple}, \ucb~\cite{jun2016anytime}, and \ns~proposed in this paper. It follows immediately from \eqref{H2M} that for $p\in(0,1)$, $\hp \leq \htwo$, and for $p\in(1,2]$, $\hp \geq \htwo$. Also, in view of \eqref{HP}, $C_p> \log K$ for $p\in(0,1)$ and $C_p< \log K$ for $p\in(1,2]$. Therefore, the comparison of $\htwo\overline{\log}~K$ and $\hp C_p$, the sample complexities of \sr~and \ns, is not obvious. As in the case of single best-arm identification, we will show that in many regimes for rewards, \ns~can outperform \sr.

Note that \ucb~\cite{jun2016anytime} is an anytime algorithm, i.e., it does not require a pre-assigned budget. In that sense, \ucb~is more powerful compared to algorithms designed specifically for the fixed-budget setting, but since it can also be used in this framework, we include it in the table as a benchmark and will compare our results with this algorithm in the numerical experiments.

\begin{figure*}
\begin{center}
\scalebox{0.91}{\fbox{
\begin{minipage}[t]{16.5cm}
\null
{\bf \large{Nonlinear Sequential Accepts and Rejects}}\\

{\bf Input:} budget $T$, parameter $p>0$.\\
\vspace{-0.24cm}

{\bf Initialize:} $\g_1=[K]$, $n_0=0$, $\m_1=M$.\\
\vspace{-0.24cm}

Let \vspace{-0.32cm}\begin{align*}
C_p&=2^{-p}+\sum_{r=2}^{K}r^{-p}\\
n_r&=\seal{\frac{T-K}{C_p{(K-r+1)}^p}} \text{for~~} r \in [K-1]
\end{align*}

At round $r=1,\ldots,K-1$:
\begin{itemize}
\item[(1)] Sample each arm in $\g_r$ for $n_r-n_{r-1}$ times. 
\item[(2)] Let $\sigma_ r: [K+1-r] \to \g_r$ be a permutation 
that orders the empirical means such that $$\mh_{\sigma_r(1),n_r} \geq \mh_{\sigma_r(2),n_r} \geq \cdots \geq \mh_{\sigma_r(K+1-r),n_r}.$$
Then, for any $\ell \in [K+1-r]$, define the following empirical gaps
\[
    \dha_{\sigma_r(\ell),n_r}= 
\begin{cases}
    \mh_{\sigma_r(\ell),n_r}-\mh_{\sigma_r(\m_r+1),n_r},& \text{if } \ell \leq \m_r\\
     \mh_{\sigma_r(\m_r),n_r}-\mh_{\sigma_r(\ell),n_r},              & \text{otherwise}
\end{cases}
\]
\item[(3)] Identify $\mathbf{index}:=\text{argmax}_{i\in \g_r} \dha_{i,n_r}$, set $\ba_r:=\{\mathbf{index}\}$ and $\g_{r+1}=\g_r \setminus \ba_r$, i.e., discard the arm $\mathbf{index}$. 
\item[(4)] If $\mh_{\mathbf{index},n_r}>\mh_{\sigma_r(\m_r+1),n_r}$, accept the arm $\mathbf{index}$, set $\m_{r+1}=\m_r-1$ and $\J_{M-\m_{r+1}}=\mathbf{index}$.
\item[(5)] After finishing $r=K-1$, the survived arm is accepted, if we have accepted $M-1$ arms at the beginning of $r=K-1$; otherwise, the survived arm is rejected.
\end{itemize}
{\bf Output:} $\{\J_1,\ldots,\J_M\}$.
\end{minipage}}}
\caption{The \ns~algorithm for identification of the best-$M$ arms.}
\label{ALGO}
\end{center}
\end{figure*}


\section{Nonlinear Sequential Accepts and Rejects}\label{Nonlinear Sequential Elimination}
In this section, we propose the Nonlinear Sequential Accepts and Rejects (\ns) algorithm for $M$-best-arm identification in the fixed budget setting. The algorithm follows the steps of \sr~\cite{bubeck2013multiple}, except for the fact that the budget allocation at each round is a nonlinear function of arms. 
The details of \ns~is given in Figure \ref{ALGO}. The algorithm is given a budget $T$ of arm pulls. At any round $r\in [K-1]$, it maintains an active set of arms $\g_r$, initialized by $\g_1=[K]$. The algorithm proceeds for $K-1$ rounds to deactivate the arms sequentially (one arm at each round) until a single arm is left. Based on an input value $p\in (0,2]$, the constant $C_p$ and the sequence $\{n_r\}_{r=1}^{K-1}$ are calculated for any $r\in [K-1]$. At round $r$, the algorithm samples the $K+1-r$ active arms for $n_r-n_{r-1}$ times and computes the empirical average of rewards for each arm. Then, it orders the empirical rewards and calculates the empirical version of gaps, where the true gaps $\Delta_i^{\left<M\right>}$ for $i\in[K]$ are defined in the first line of \eqref{H2M}. 
The arm with the highest empirical gap is deactivated: if its empirical reward is within the top $M$ arms, it is accepted; otherwise, it is rejected. At the end, the algorithm outputs $M$ accepted arms as the top $M$ arms. 

Note that our algorithm with the choice of $p=1$ amounts to \sr. We will show that in many regimes for arm gaps, $p\neq 1$ provides better theoretical results, and we further exhibit the efficiency in the numerical experiments in Section \ref{simul}. The following proposition encapsulates the theoretical guarantee of the algorithm (the proof is given in the appendix).
\begin{proposition}\label{mainprop}
{\it Let the Nonlinear Sequential Accepts and Rejects algorithm in Figure \ref{ALGO} run for a given $p\in (0,2]$, and let $C_p$ and $\hp$ be defined as in \eqref{HP} and \eqref{H2M}. Then, the misidentification probability satisfies the bound, }
$$
\p{\{\J_1,\ldots,\J_M\} \neq [M]} \leq 2K^2\ex{-\frac{T-K}{8C_p\hp}}.
$$ 
\end{proposition}
The performance of \ns~relies on the input parameter $p$, but this choice is more qualitative rather than quantitative. In particular, larger values for $p$ increase $\hp$ and decrease $C_p$, and hence, there is a trade-off in selecting $p$. According to Table \ref{table20}, to compare \ns~with \sr~and \ucb~, we have to evaluate the corresponding sample complexities. Fair theoretical comparisons with \ucb~is delicate, since $\hone$ is in essence slightly different from $\htwo$ and $\hp$. However, we will provide comprehensive simulations in Section \ref{simul} to compare all algorithms. We consider two instances for sub-optimality of arms in this section to compare \ns~with \sr:
\begin{itemize}
\item[{\bf 1}] {\bf A large group of competitive arms:} The top $M$ arms are roughly similar such that $\mu_1 \approx \mu_M$, $\mu_M-\mu_{M+1}=\delta_1$ is non-negligible, and the other arms are just as competitive as each other, i.e., $\mu_{M+1} \approx \mu_K$.
\item[{\bf 2}] {\bf A small group of competitive arms:} The top $M$ arms are roughly similar such that $\mu_1 \approx \mu_M$. $\mu_M-\mu_{M'}=\delta_1$ for a small number of arms ($M'=\oo(1)$ with respect to $K$) and $\mu_{M+1}\approx \mu_{M'}$, $\mu_{M'}-\mu_{M'+1}=\delta_2$, and $\mu_{M'+1}\approx \mu_{K}$. We also have $\delta_1 \ll \delta_2$.
\end{itemize}
The subsequent corollary follows from Proposition \ref{mainprop}. Note that the orders are expressed with respect to $K$.
\begin{corollary}\label{ratecorollary}
{\it Consider the Nonlinear Sequential Accepts and Rejects algorithm in Figure \ref{ALGO}. Let constants $p$ and $q$ be chosen such that $1<p\leq2$ and $0<q<1$. Then, for the two settings given above, the bound on the misidentification probability presented in Proposition \ref{mainprop} satisfies}
\begin{table}[h!]
\begin{center}
\scalebox{1.15}{
\begin{tabular}{ | c | c |} 
 \hline 
   {\bf Regime 1} &    {\bf Regime 2} \\ [.5 ex]
 \hline 
 $C_qH^{\left<M\right>}(q)=\oo(K)$ & $C_p\hp=\oo(1)$ \\ [.5 ex]
 \hline
\end{tabular}}
\end{center}
\end{table}
\end{corollary} 
Now let us compare \ns~and \sr~using the result of Corollary \ref{ratecorollary}. Returning to Table \ref{table20} and calculating $\htwo$ for Regimes $1$ and $2$, we can derive the following table, 
\begin{table}[h!]
\caption{The sampling complexity for \ns~(our algorithm) and \sr. For Regime 1, we set $0<q<1$, and for Regime 2, we use $1< p \leq 2$. The order does not include the $\log \frac{K}{\delta}$ term as it is in common between the two algorithms.}
\begin{center}
\scalebox{1.25}{
\begin{tabular}{| c | c | c  |} 
 \hline 
 Algorithm & \sr &  \ns \\ [.5 ex]
 \hline 
 {\bf Regime 1} & $\oo(K\log K)$ & $\oo(K)$ \\ [.5 ex]
 \hline
{\bf Regime 2} &  $\oo(\log K)$ & $\oo(1)$\\ [.5 ex]
 \hline
\end{tabular}}\label{table2}
\end{center}
\end{table}
which shows that with a proper tuning for $p$, we can save a $\mathcal{O}(\log K)$ factor in the sampling complexity. Though we do not have prior information on gaps to categorize them specifically, the choice of the input parameter $p$ is more qualitative rather than quantitative, i.e., once the sub-optimal arms are almost the same $0<p<1$ performs better than $1<p\leq2$, and when there are a few real competitive arms, $1<p\leq2$ outperforms $0<p<1$. Next, we will show in the numerical experiments that a wide range of values for $p$ can potentially result in efficient algorithms with small misidentification error. 


\section{Numerical Experiments}\label{simul}

\begin{figure*}[t!]
\begin{center}
        \includegraphics[scale=.7]{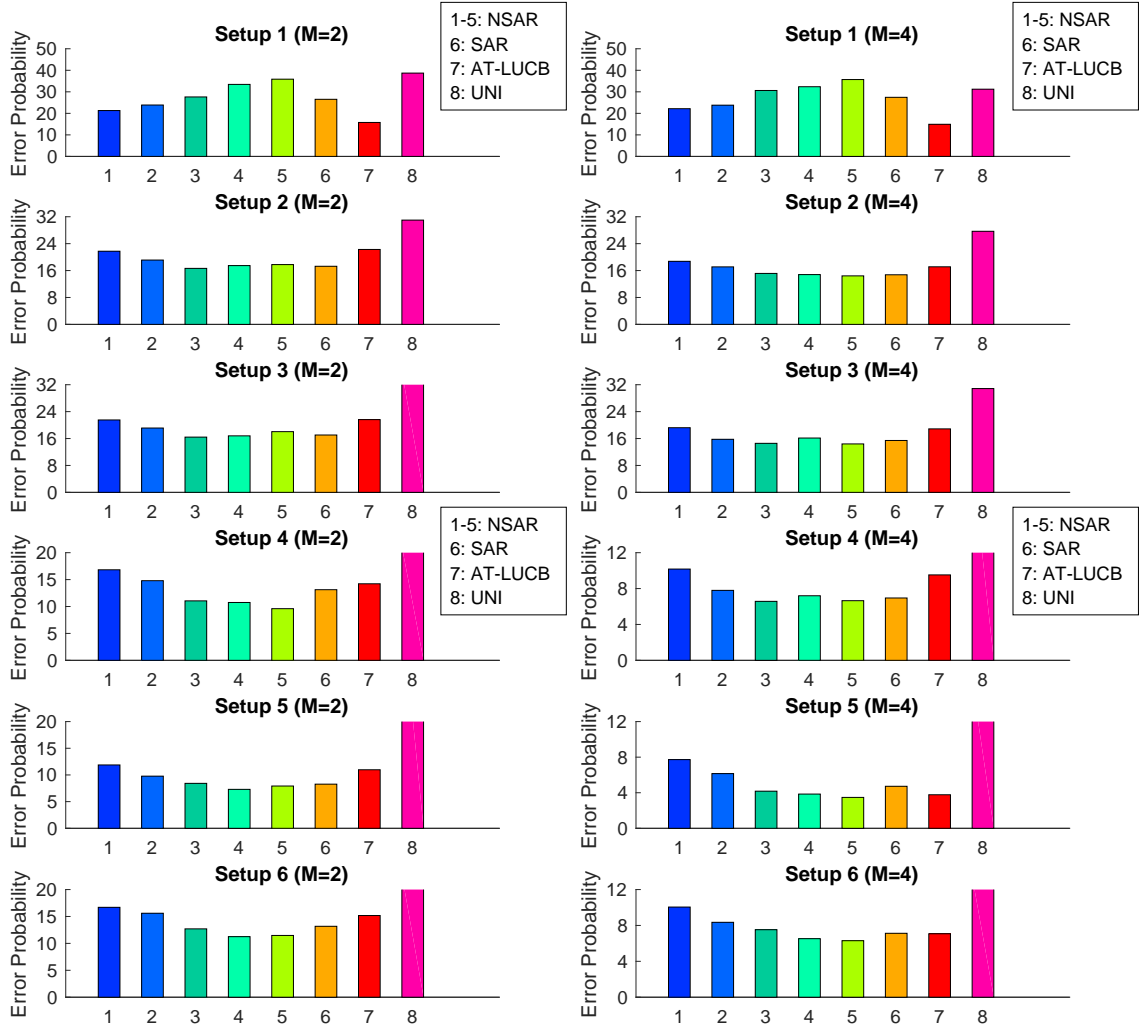}
\end{center}
\vspace{-1.79cm}
	\caption{The figure shows the misidentification probability for \ns, \sr, \ucb, and \uni~algorithms in six different setups. The six plots on the left relate to the case $M=2$, and the six plots on the right are associated with $M=4$. The height of each bar represents the misidentification probability, and each index (or color) represents one algorithm tuned with a specific parameter.}
	\label{plot}
\end{figure*}

We now empirically evaluate our proposed algorithm on a few settings studied in \cite{bubeck2013multiple}. More specifically, we compare \ns~with \sr, \ucb, as well as uniform allocation (\uni), where in the \uni~algorithm, we simply divide the budget uniformly across the arms. We remark that \ucb~in \cite{jun2016anytime} is an anytime algorithm, i.e., it does not require a pre-assigned budget; however, since it can also be used for the fixed-budget setting, we include it in our numerical experiments as a benchmark. We consider $K=50$ arms and assume Bernoulli distribution on the rewards. For the following setups, we examine two values for top arms $M \in \{2,4\}$ (we use the notation $x\cl{y}$ to denote integers in $[x,y]$):
\begin{itemize}
\item[{\bf 1}]{\bf One group of suboptimal arms:} $\mu_{1:M}=0.7$ and $\mu_{M+1:K}=0.5$.
\item[{\bf 2}]{\bf Two groups of suboptimal arms:} $\mu_{1:M}=0.7$, $\mu_{M+1:2M}=0.66$, and $\mu_{2M+1:K}=0.5$.
\item[{\bf 3}]{\bf Three groups of suboptimal arms:} $\mu_{1:M}=0.7$, $\mu_{M+1:2M}=0.66$, $\mu_{2M+1:3M}=0.62$, and $\mu_{3M+1:K}=0.5$.
\item[{\bf 4}]{\bf Beta(2,2):} The expected values of Bernoulli distributions are generated according to a beta distribution with shape parameters $2$ and $2$.
\item[{\bf 5}]{\bf Beta(5,5):} The expected values of Bernoulli distributions are generated according to a beta distribution with shape parameters $5$ and $5$.
\item[{\bf 6}]{\bf One real competitive arm:} $\mu_{1:M}=0.7$, $\mu_{M+1}=0.68$ and $\mu_{M+2:K}=0.5$.
\end{itemize}
We run $4000$ experiments for each setup with a specific value of $M$, and we calculate the misidentification probability by averaging out over the error in experiment runs. We set the budget $T$ in each setup equal to $\seal{\hone}$ in the corresponding setup as suggested in \cite{bubeck2013multiple}, and we also choose the parameters of \ucb~as instructed in \cite{jun2016anytime}.

We illustrate the overall performance of the algorithms in Figure \ref{plot} for different setups. The height of each bar shows the misidentification probability, and the index guideline is as follows: {\bf (i)} indices 1-5: \ns~with parameter $p\in \{0.7,0.85,1.1,1.2,1.3\}$. {\bf (ii)} index 6: \sr. {\bf (iii)} index 7: \ucb. {\bf (iv)} index 8: \uni. The legends are the same for all of the plots, and hence, they are omitted in most of the plots.

The results are consistent with Corollary \ref{ratecorollary}, and the following comments are in order:
\begin{itemize}
\item Setup 1 corresponds to Regime 1 in Corollary \ref{ratecorollary}. As expected, with any choice of $0<p<1$, \ns~should outperform \sr, and we observe that this happens when $p\in \{0.7,0.85\}$. However, in this regime, our algorithm is inferior compared to \ucb.
\item Setups 2-3-6 are considered close to Regime 2 in Corollary \ref{ratecorollary} as we have a small number of arms competitive to the top $M$ arms. Thus, we should choose $1<p\leq 2$. We observe that in these setups, at least for two choices out of $p\in \{1.1,1.2,1.3\}$, \ns~outperforms \sr~and \ucb. One should observe that the improvement in Corollary \ref{ratecorollary} is $\oo(\log K)$ which increases slowly with $K$. Since we only have $K=50$ numbers, using larger values for $p$ is not suitable in these setups, because the increase in $\hp$ worsens the performance overall. Though for larger values of $K$, the improvement must be more visible, we avoid that due to prohibitive time-complexity of Monte Carlo simulations. 
\item In Setups 4-5, we choose the expected values of Bernoulli rewards randomly and concentrate them around $0.5$. Again, for at least two choices out of $p\in \{1.1,1.2,1.3\}$, our algorithm outperforms \sr~and \ucb.
\item In all setups, the naive \uni~algorithm is outperformed by the other methods.
\end{itemize}

Overall, the performance of algorithms depends on the problem environment. If we have prior knowledge of the environment, we can select the suitable algorithm. The notable feature of \ns~is incorporation of this prior knowledge in tuning of $p$ without changing the foundation of the algorithm.

\section{Conclusion}\label{Conclusion}
We considered $M$-best-arm identification in stochastic multi-armed bandits, where the objective is to find the top $M$ arms in the sense of the expected reward. We presented an algorithm working based on sequential deactivation of arms in rounds. The key is to allocate the budget of arm pulls in a nonlinear fashion at each round. We proved theoretically and empirically that we can gain from the nonlinear budget allocation in several problem environments, compared to the state-of-the-art methods. An important future direction is to propose a method that adaptively fine-tunes the nonlinearity according to the problem environment.


\section{Appendix}\label{Appendix}
\begin{fact}(Hoeffding's inequality) \label{HOFF}
{\it Let $W_1,\ldots,W_n$ be independent random variables with support on the unit interval with probability
one. If $S_n=\sum_{i=1}^nW_i$, then for all $a>0$, it holds that
$$
\p{S_n-\E[S_n]\geq a} \leq \ex{\frac{-2a^2}{n}}. 
$$}
\end{fact}

\section*{Proof of Proposition \ref{mainprop}}
Recall that $\mh_{i,n}$ denotes the average reward of pulling arm $i$ for $n$ times. Now consider the following event 

\scalebox{.9}{\parbox{\linewidth}{%
\begin{align*} 
\mathcal{E}:=\left\{\forall i \in [K], \forall r\in [K-1]: \abs{\mh_{i,n_k}-\mu_i} \leq \frac{1}{4}\Delta^{\left<M\right>}_{(K+1-r)}\right\}.
\end{align*}}}

Using Hoeffding's inequality (Fact \ref{HOFF}), we get

\begin{align*}
\p{\mathcal{E}^C} &\leq \sum_{i=1}^{K}\sum_{r=1}^{K-1} \p{\abs{\mh_{i,n_k}-\mu_i} > \frac{1}{4}\Delta^{\left<M\right>}_{(K+1-r)}} \\  
&\leq  \sum_{i=1}^{K}\sum_{r=1}^{K-1} 2 \ex{-2n_r\left(\frac{1}{4}\Delta^{\left<M\right>}_{(K+1-r)}\right)^2}.
\end{align*}
Noting the fact that $n_r=\seal{\frac{T-K}{C_p(K+1-r)^p}} \geq \frac{T-K}{C_p(K+1-r)^p}$, we can use above to conclude that

\scalebox{.9}{\parbox{\linewidth}{%
\begin{align*}
\p{\mathcal{E}^C}& \leq 2K^2\max_{r\in [K-1]}\left\{\ex{-\frac{T-K}{8}\frac{\left(\Delta^{\left<M\right>}_{(K+1-r)}\right)^2}{C_p(K+1-r)^p}}\right\}\\
&= 2K^2\ex{-\frac{T-K}{8}\min_{r\in [K-1]}\left\{\frac{\left(\Delta^{\left<M\right>}_{(K+1-r)}\right)^2}{C_p(K+1-r)^p}\right\}}\\
&= 2K^2\ex{-\frac{T-K}{8C_p\hp}}.
\end{align*}}}
The rest of the proof is to show that the event $\mathcal{E}$ warrants that the algorithm does not make erroneous decision. This part follows precisely by the induction argument given in \cite{bubeck2013multiple} (see page 4-5). \hfill $ \square$

\section*{Proof of Corollary \ref{ratecorollary}}
First, let us analyze the order of $C_p$ defined as 
$$
C_p=2^{-p}+\sum_{r=2}^{K}r^{-p}.
$$
For any $p>1$, $C_p$ is a convergent sum when $K\rightarrow \infty$. Thus, for the regime $p>1$, the sum is a constant, i.e., $C_p=\mathcal{O}(1)$. On the other hand, consider $q \in (0,1)$, and note that the sum is divergent, and for large $K$ we have $C_q = \mathcal{O}(K^{1-q})$. Now, let us analyze
$$
\hp=\max_{i\neq 1}\left\{i^p\left(\Delta_{(i)}^{\left<M\right>}\right)^{-2}\right\}.
$$
For Regime 1, $q \in (0,1)$ and we have
$$
\max_{i\neq 1}\left\{i^q\left(\Delta_{(i)}^{\left<M\right>}\right)^{-2}\right\} \approx \frac{K^q}{\delta_1^2}
$$
Combining with $C_q$, the product $C_qH^{\left<M\right>}(q)=\mathcal{O}(K)$. For Regime 2, $p\in(1,2]$ and we have 
$$
\max_{i\neq 1}\left\{i^p\left(\Delta_{(i)}^{\left<M\right>}\right)^{-2}\right\} \approx \frac{{M'}^p}{\delta_1^2}=\oo(1),
$$
since $\delta_1 \ll \delta_2$. 
Therefore, combining with $C_p=\mathcal{O}(1)$, the product $C_p\hp=\mathcal{O}(1)$.
\hfill $ \square$

\vspace{.9cm}
\bibliographystyle{IEEEtran}
\bibliography{IEEEabrv,references}


\end{document}